\pdfoutput=1

\documentclass[11pt]{article}

\usepackage[]{ACL2023}

\usepackage{times}
\usepackage{latexsym}

\usepackage[T1]{fontenc}

\usepackage[utf8]{inputenc}

\usepackage{microtype}

\usepackage{inconsolata}

\usepackage{booktabs}
\usepackage{graphicx}
\usepackage{siunitx}
\usepackage{amsmath}

\usepackage{todonotes}

\title{Progressive Translation: Improving Domain Robustness of Neural Machine Translation with Intermediate Sequences\thanks{\ \ The work described in this paper is substantially supported by a grant from the Research Grant Council of the Hong Kong Special Administrative Region, China (Project Code: 14200620).}}

\author{Chaojun Wang$^1$ \quad Yang Liu$^{2}$ \quad Wai Lam$^{1}$ \bigskip\\
  $^1$The Chinese University of Hong Kong \\
  $^2$Microsoft Cognitive Services Research \\
  \texttt{cj.wang@link.cuhk.edu.hk}}

\begin{document}
\maketitle
\begin{abstract}
Previous studies show that intermediate supervision signals benefit various Natural Language Processing tasks. However, it is not clear whether there exist intermediate signals that benefit Neural Machine Translation (NMT). Borrowing techniques from Statistical Machine Translation, we propose intermediate signals which are intermediate sequences from the "source-like" structure to the "target-like" structure. 
Such intermediate sequences introduce an inductive bias that reflects a domain-agnostic principle of translation, which reduces spurious correlations that are harmful to out-of-domain generalisation.
Furthermore, we introduce a full-permutation multi-task learning to alleviate the spurious causal relations from intermediate sequences to the target, which results from \textit{exposure bias}. The Minimum Bayes Risk decoding algorithm is used to pick the best candidate translation from all permutations to further improve the performance. Experiments show that the introduced intermediate signals can effectively improve the domain robustness of NMT and reduces the amount of hallucinations on out-of-domain translation. Further analysis shows that our methods are especially promising in low-resource scenarios.

\end{abstract}

\section{Introduction}

A spectrum of studies recently arose in Natural Language Processing (NLP), which incorporates intermediate supervision signals into the model by simply converting the intermediate signals into textual sequences and prepending or appending these sequences to the output sequence. It benefits tasks such as math word problems~\citep{wei2022chain}, commonsense reasoning~\citep{liu-etal-2022-generated}, programs execution~\citep{nye2022show},  summarisation~\citep{narayan-etal-2021-planning}, etc. This trend further triggered the collection of a new dataset with intermediate results~\citep{lewkowycz2022solving} and corresponding theoretical analysis~\citep{wies2022subtask}. Intermediate supervision signals show consistent benefits to these various sequence generation tasks and Neural Machine Translation (NMT) is a basic and typical sequence generation task in the NLP community. 
However, it remains an open question whether and how intermediate signals can be defined and leveraged for NMT.

Meanwhile, previous studies~\citep{koehn-knowles-2017-six, muller-etal-2020-domain} found that NMT suffers from poor domain robustness, i.e. the generalisation ability to unseen domains. Such an ability not only has theoretical meaning, but also has practical value since: 1) the target domain(s) may be unknown when a system is built; 2) some language pairs may only have training data for limited domains. Since the recent study~\citep{wei2022chain} in intermediate supervision signals showed a benefit of such signals on out-of-domain generalisation, we expect intermediate signals may benefit domain robustness in NMT. 

\begin{figure}[!t]
    \centering
    \includegraphics[width=7.5cm]{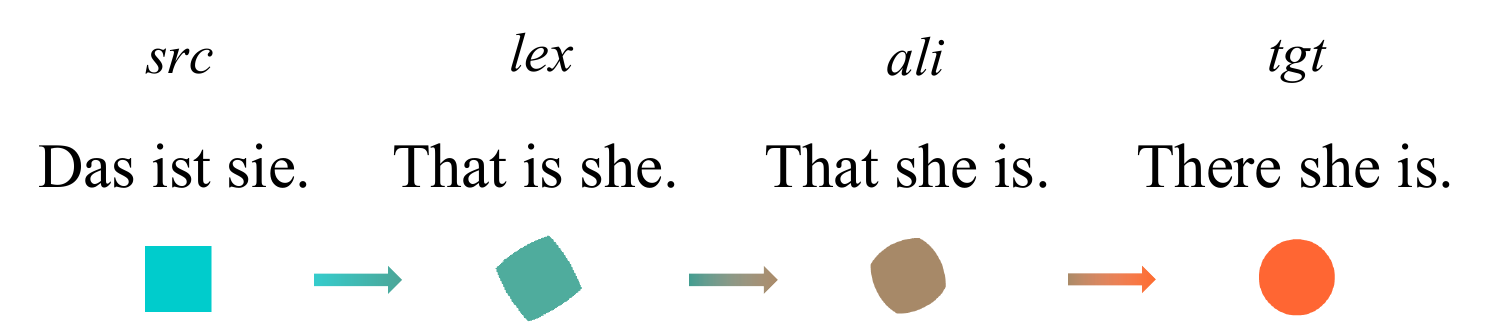}
    \caption{An illustration of the transformation from a source sentence to the target translation and its analogy with vision. \textit{src}: source; \textit{tgt}: target; \textit{lex}: word-by-word translation; \textit{ali}: reorders \textit{lex} monotonically based on word alignments.}
    \label{fig:1}
\end{figure}

Different from math problem-solving tasks, machine translation tasks do not have explicit intermediate results to serve as the intermediate signals. A recent work~\citep{voita-etal-2021-language} found that NMT acquires the three core SMT competencies, target-side language modelling, lexical translation and reordering in order during the course of the training. Inspired by this work, we borrow techniques in SMT to produce intermediate sequences as the intermediate signals for NMT. Specifically, we first obtain the word alignments for the parallel corpus and use it to produce the word-for-word translations (\textit{lex}) and the aligned word-for-word translations (\textit{ali}) to resemble the lexical translation and reordering competencies in SMT. As shown in Figure~\ref{fig:1}, the intermediate sequences resemble structurally approaching the target from the source progressively, which shares a similar spirit of how humans do translation or reasoning about translation step by step, thus named Progressive Translation.

Our intuition is that these intermediate sequences inject an inductive bias about a domain-agnostic principle of the transformation between two languages, i.e. word-for-word mapping, then reordering, and finally refinement. Such a bias limits the learning flexibility of the model but prevents the model from building up some spurious correlations~\citep{arjovsky2019invariant} which harm out-of-domain performance.

However, previous works have shown that NMT is prone to overly relying on the target history~\citep{wang-sennrich-2020-exposure, voita-etal-2021-analyzing}, which is partially correlated with \textit{exposure bias}~\citep{DBLP:journals/corr/RanzatoCAZ15} (a mismatch between training and inference), especially under domain-shift. Simply prepending these introduced intermediate sequences to the target would introduce spurious causal relationships from the intermediate sequences to the target. As a result, these intermediate sequences would potentially mislead the model about the prediction of the target, due to erroneous intermediate sequences during inference. To alleviate this spurious causal relationship, we introduce the full-permutation multi-task learning framework, where the target and intermediate sequences are fully permuted. The Minimum Bayes Risk~\citep{GOEL2000115} decoding algorithm is used to select a \textit{consensus} translation from all permutations to further improve the performance.

We first test our proposed framework on IWSLT'14 German$\to$English and find that the proposed intermediate sequence can improve the domain robustness of NMT. The permutation multi-task learning is important for the intermediate sequence which is prone to erroneous during inference. To examine the generality of our methods, we conduct experiments on another two domain-robustness datasets in NMT, OPUS German$\to$English and a low resource German$\to$Romansh scenario. Our methods show consistent out-of-domain improvement over these two datasets.

Moreover, previous works~\citep{muller-etal-2020-domain, wang-sennrich-2020-exposure} found that hallucinated translations are more pronounced in out-of-domain setting. Such translations are fluent but completely unrelated to the input, and they may cause more serious problems in practical use due to their misleading nature. Therefore, we manually evaluate the proportion of hallucinations. Results show that our methods substantially reduce the amount of hallucinations in out-of-domain translation. Finally, since the corpus size in the main experiments is relatively small, we investigate the effectiveness of our methods when scaling up the corpus sizes. Results show that our methods are especially effective under the low-resource scenarios.

\begin{figure*}[!t]
    \centering
    \includegraphics[width=0.99\linewidth]{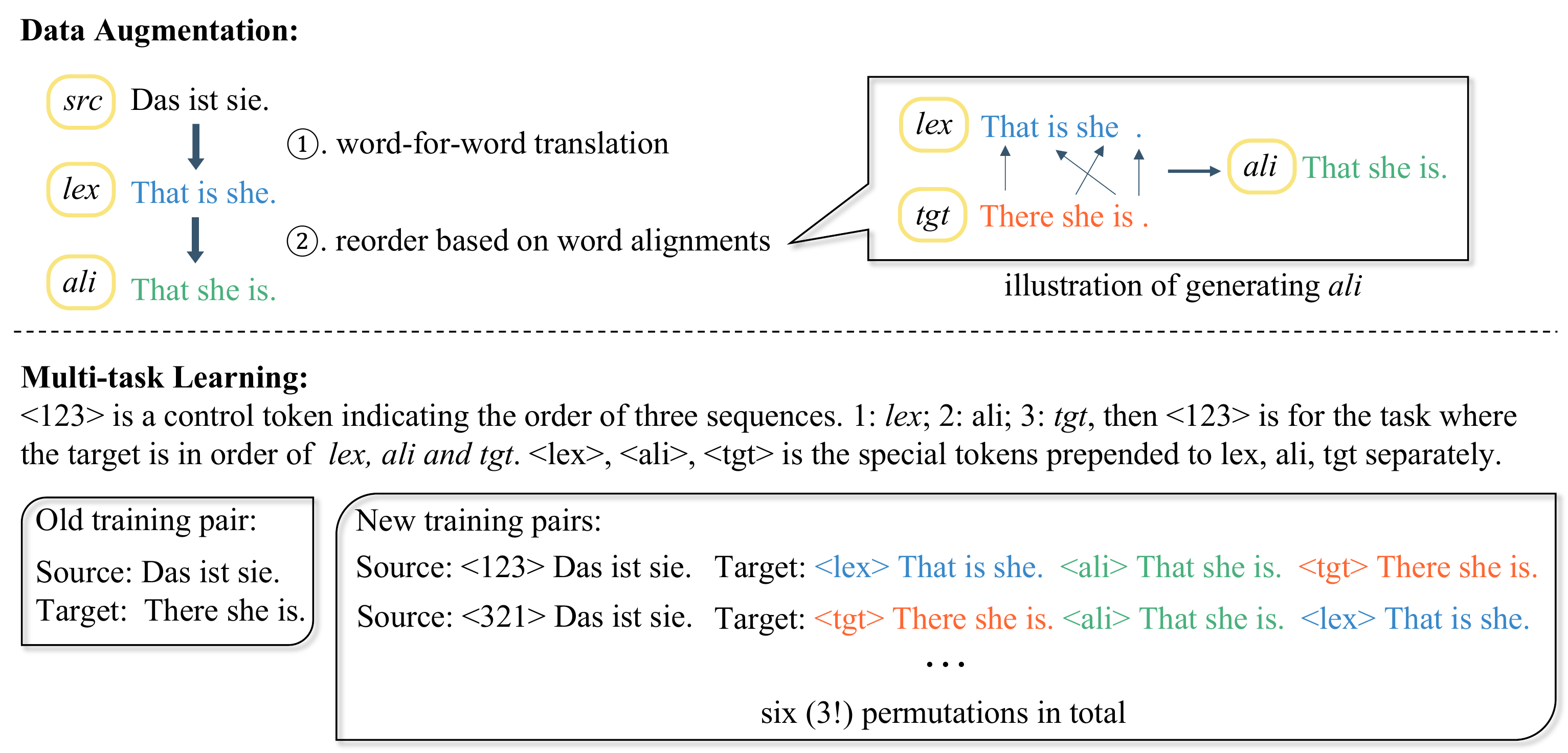}
    \caption{An illustration of the proposed intermediate sequences and multi-task learning framework. src: source.}
    \label{fig:3}
\end{figure*}

\section{Related Work}
\textbf{Intermediate Supervision Signals.}
Some existing works in the broader NLP community try to incorporate intermediate sequences into the model. We take two typical examples of them to better distinguish our work from other works. \citet{narayan-etal-2021-planning} uses an entity chain as the intermediate sequence for summarisation. \citet{wei2022chain} produces intermediate sequences resembling the deliberation process of humans. Similar to~\citet{narayan-etal-2021-planning}, Progressive Translation (PT) augments data for the whole training set and the intermediate sequences are not limited to literally understandable sequences. Similar to~\citet{wei2022chain}, sequences augmented by PT resemble approaching the output from the input.

\noindent\textbf{Data Augmentation of Domain Robustness in NMT.}
Existing works in data augmentation try to improve the domain robustness of NMT by introducing more diverse synthetic training examples~\citep{ng-etal-2020-ssmba} or auxiliary tasks where the target history is less informative~\citep{sanchez-cartagena-etal-2021-rethinking} named MTL-DA framework. 
The main difference between our PT framework and the MTL-DA framework is that the MTL-DA framework treats each target-side sequence as an independent task conditioned on the source, whereas PT also encourages the model to learn the transformational relations between any pair of target-side sequences, which may help the model to generalise better across domains.

\noindent\textbf{Statistical Machine Translation in NMT.}
The intermediate sequences of PT are produced using the word alignments and reordering components in Statistical Machine Translation (SMT). There are works on improving NMT with SMT features and techniques~\citep{He_He_Wu_Wang_2016, chen-etal-2016-guided, du2017pre, zhao-etal-2018-exploiting}. However, these works either modify the architecture of the neural network or require more than one model to produce the translation (e.g. a rule-based pre-ordering model and a NMT model etc.). To the best of our knowledge, we are the first to incorporate features from SMT into NMT by converting the features into textual sequences and prepending these to the target without requiring extra models or modifying the neural architecture.

\section{Approach}

\subsection{Intermediate Sequences}
The traditional SMT decomposes the translation task into distinct components where some features could potentially be the intermediate supervision signals. More recently, \citet{voita-etal-2021-language} found that NMT acquires the three core SMT competencies, i.e. target-side language modelling, lexical translation and reordering, in order during the course of training. Inspired by this work, we produce word-for-word translations and aligned word-for-word translations as the intermediate sequences to resemble the lexical translation and reordering components separately using the word alignments component in SMT.

As shown in Figure~\ref{fig:3} Data Augmentation part, for each source-target parallel sequence in the training corpus, we augment their target sequences with two extra intermediate sequences, \textit{lex} and \textit{ali}. The two intermediate sequences are prepended to the target to form an augmented target.

\textbf{lex}: The source sequence is word-for-word translated based on a bilingual lexicon obtained from the parallel training corpus. Tokens that are not in the lexicon are copied into \textit{lex}.

\textbf{ali}: \textit{lex} is reordered so that the word alignments from the target to \textit{lex} is monotonic. The word alignments used here are target-to-source alignments because it is equivalent to the target-to-\textit{lex} alignments since \textit{lex} is word-for-word mapped from the source. The words in the target which is assigned to "NULL" are omitted during reordering. 

\textit{lex}, \textit{ali} and target (\textit{tgt}) are prefixed with a special token separately for extracting the corresponding sequence from the predicted output. The one-to-many (both source-to-target and target-to-source) word alignments are obtained with \textit{mgiza++}~\citep{gao-vogel-2008-parallel, och-ney-2003-systematic}\footnote{\url{https://github.com/moses-smt/mgiza}}, a SMT word alignments tool, on the \textbf{in-domain} training corpus, following the default parameter provided in \textit{train-model.perl} by Moses~\citep{koehn-etal-2007-moses}\footnote{\url{https://github.com/moses-smt/mosesdecoder/blob/master/scripts/training/train-model.perl}}. The one-to-one word alignments are built by computing the intersection between the one-to-many word alignments in both directions. The bilingual lexicon is obtained by associating each source word to the target word it is most frequently aligned within the one-to-one word alignments.

The learning of word alignments and transformations of \textit{lex} and \textit{ali} are at the word level. The BPE~\citep{sennrich-etal-2016-neural} word segmentation is trained on \textit{src}-\textit{tgt} parallel data as normal and applied to both source-target parallel sequences and intermediate sequences (the target-language vocabulary is applied to split the words in the intermediate sequences). 

We expect that the introduced intermediate sequences would benefit the domain robustness of NMT. Because the proposed intermediate sequences serve as a supervision signal to provide the model with an explicit path for learning the transformational relations from source to target. Such signals inject an inductive bias about one kind of domain-agnostic principle of the transformation between two languages, i.e. word-for-word mapping, then reordering, finally refinement. This injected bias limits the learning flexibility of the neural model but prevents the model from building up some spurious correlations which harm out-of-domain performance.

\begin{figure}[!t]
    \centering
    \includegraphics[width=6cm]{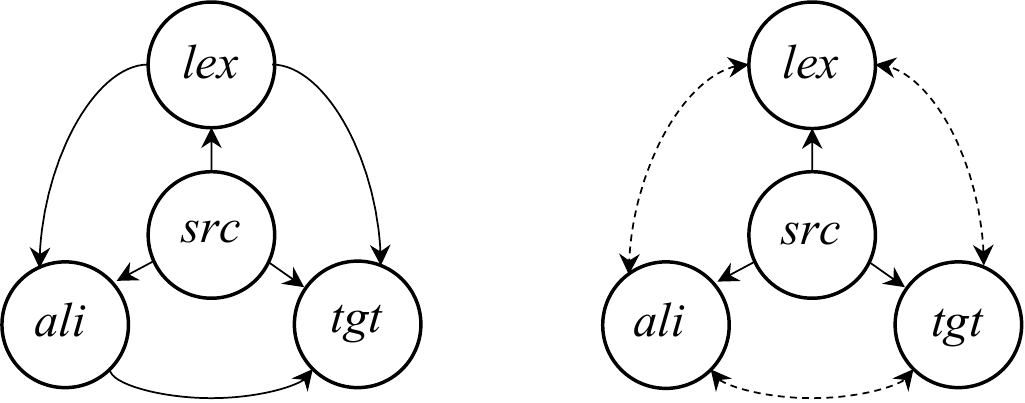}
    \caption{Causal graphs for the source and three target-side sequences. Solid arrow denotes casual dependence and dashed arrow represents the statistical correlation between two variables. Left: relations if we simply prepend \textit{lex} and \textit{ali} to the target. Right: relations after full-permutation multi-task learning.}
    \label{fig:2}
\end{figure}

\subsection{Spurious Causality Relationship}

To introduce these intermediate sequences as intermediate supervision signals to the model, we prepend them to the output sequence in training. However, simply prepending these produced intermediate sequences to the target would potentially introduce spurious causality relationships from pre-sequence to post-sequence. For example, prepending \textit{lex}, \textit{ali} to the target would introduce the causal relationships of \textit{lex} $\to$ \textit{ali} $\to$ \textit{tgt}. These are spurious causality relationships because the model is highly unlikely to get the gold-standard pre-sequences (\textit{lex} or \textit{ali}) as in the training during inference, especially under the domain-shift where the performance is relatively poor. Therefore, the model should learn that source (input) is the only reliable information for any target-side sequences. Note that such spurious causality relationship in principle results from a mismatch between training and inference of the standard training-inference paradigm of NMT, which is termed \textit{exposure bias} by the community.

Intuitively, if the model could predict the target-side sequences in any order, then the causality relationship between target-side sequences should be reduced. Therefore, we propose to fully permute the target-side sequences, i.e. intermediate sequences (\textit{lex} or \textit{ali}) and the target sequence (\textit{tgt}). Figrue~\ref{fig:3} illustrates the training data after permutation when we prepend both \textit{lex} and \textit{ali} to the target. The source is prefixed with a control token for each permutation, i.e. 1: \textit{lex}; 2: \textit{ali}; 3: \textit{tgt}, then <123> is the control token for the permutation where the target is in the order of \textit{lex}, \textit{ali} and \textit{tgt}.

As shown in Figure~\ref{fig:2}, with the permutation, we create counterfactual data which disentangles the causal relations of \textit{lex} $\to$ \textit{ali} $\to$ \textit{tgt} and enhances the causal relations from source to each of these three sequences. Therefore, the full-permutation multi-task training better balances the model's reliance on the source and target history, at least on pre-sequence(s).

\subsection{Minimum Bayes Risk Decoding}
From our preliminary experiments, we found that various test sets prefer different generation orders of the permutation. For example, order \textit{lex-ali-tgt} performs best on some test sets whereas \textit{tgt-ali-lex} performs best on some other test sets. Therefore, we suspect that the translation quality would be further improved if we could dynamically select the best candidate translations from all permutations. Inspired by~\citep{eikema2021samplingbased}, we use Minimum Bayes Risk (MBR) decoding to select a \textit{consensus} translation from all permutations.

\begin{table*}[!t]
\centering
\begin{small}
\begin{tabular}{l|l|ccccc}
\toprule
\textbf{ID} & \textbf{Augmentation} & \textbf{In-Domain} &\textbf{IT} & \textbf{Law} & \textbf{Medical} & \textbf{average OOD} \\
\midrule
1 & Transformer & 32.1$_{\pm 0.38}$ & 14.7$_{\pm 0.21}$ & 10.1$_{\pm 0.38}$ & 17.0$_{\pm 0.25}$ & 13.9$_{\pm 0.19}$ \\
\midrule
2 & \textit{lex}+\textit{tgt} & 31.2$_{\pm 0.50}$ & 16.6$_{\pm 0.26}$ & 11.1$_{\pm 0.23}$ & 20.7$_{\pm 0.66}$ & 16.1$_{\pm 0.30}$ \\
3 & \textit{ali}+\textit{tgt} & 25.8$_{\pm 3.57}$ & 14.4$_{\pm 2.54}$ & \phantom{0}4.5$_{\pm 6.00}$ & 17.9$_{\pm 1.32}$ & 12.2$_{\pm 3.25}$ \\
4 & \textit{lex}+\textit{ali}+\textit{tgt} & 25.5$_{\pm 7.82}$ & \phantom{0}9.4$_{\pm 1.14}$ & \phantom{0}3.1$_{\pm 2.31}$ & 11.3$_{\pm 6.70}$ & \phantom{0}7.9$_{\pm 1.71}$ \\
\midrule
5 & 2 + permu & 30.1$_{\pm 1.55}$ & 15.5$_{\pm 0.50}$ & \phantom{0}7.2$_{\pm 5.48}$ & 19.0$_{\pm 1.08}$ & 13.9$_{\pm 2.18}$ \\
6 & 3 + permu & 30.6$_{\pm 0.30}$ & 16.9$_{\pm 1.00}$ & 10.8$_{\pm 0.40}$ & 19.9$_{\pm 0.60}$ & 15.9$_{\pm 0.53}$ \\
7 & 4 + permu & 29.9$_{\pm 0.32}$ & 18.2$_{\pm 0.89}$ & 10.8$_{\pm 0.10}$ & 20.7$_{\pm 0.40}$ & 16.6$_{\pm 0.37}$ \\
\midrule
8 & 7 + MBR & 30.5$_{\pm 0.21}$ & 17.7$_{\pm 0.72}$ & 11.8$_{\pm 0.1}$ & 21.6$_{\pm 0.49}$ & 17.0$_{\pm 0.35}$ \\
\bottomrule
\end{tabular}
\end{small}
\caption{Average BLEU ($\uparrow$) and standard deviation of ablation results on in-domain and out-of-domain test sets on IWSLT'14 DE$\to$EN. permu: permutation.}
\label{tab:6}
\end{table*}

MBR aims to find a translation that maximises expected utility (or minimises expected risk) over the posterior distribution. In practice, the posterior distribution is approximated by drawing a pool of samples $\mathcal{S} = (s_1,...,s_n)$ of size $n$ from the model:
\begin{equation}
\label{eq:1}
y^{\star}=\underset{s_{i} \in \mathcal{S}}{\operatorname{argmax}} \frac{1}{n} \sum_{s_{j}=1}^{n} u\left(s_{i}, s_{j}\right)
\end{equation}
where $u$ is the utility function to compute the similarity between two sequences. In our experiment, the samples $\mathcal{S}$ are translations from all permutations.

Following~\citet{eikema2021samplingbased}, we use BEER~\citep{stanojevic-simaan-2014-fitting} as the utility function, and the released toolkit\footnote{\url{https://github.com/Roxot/mbr-nmt}} for MBR decoding.

\section{Experiments}
\subsection{Dataset}
We work on three datasets involving two language pairs, which were used in previous works on the domain robustness in NMT~\citep{sanchez-cartagena-etal-2021-rethinking, ng-etal-2020-ssmba}.

\textit{IWSLT'14 DE$\to$EN} IWSLT'14~\citep{cettolo-etal-2014-report} German$\to$English (DE$\to$EN) is a commonly used small-scale dataset in NMT, which consists of \num{180000} sentence pairs in the TED talk domain. Following~\citet{sanchez-cartagena-etal-2021-rethinking}, the validation and in-domain (ID) testing sets are \textit{tst2013} and \textit{tst2014} separately; and out-of-domain (OOD) test sets consist of \textit{IT}, \textit{law} and \textit{medical} domains from OPUS~\citep{lison-tiedemann-2016-opensubtitles2016} collected by~\citet{muller-etal-2020-domain}\footnote{\url{https://github.com/ZurichNLP/domain-robustness}}.

\textit{OPUS DE$\to$EN \& Allegra DE$\to$RM}  are two benchmarks of domain-robustness NMT  released by~\citet{muller-etal-2020-domain}. OPUS comprises  five domains: \textit{medical}, \textit{IT}, \textit{law}, \textit{koran} and \textit{subtitles}. Following~\citet{ng-etal-2020-ssmba}, we use \textit{medical} as ID for training (which consists of \num{600000} parallel sentences) and validation and the rest of four domains as OOD test sets. Allegra~\citep{scherrer-cartoni-2012-trilingual} German$\to$Romansh (DE$\to$RM) has \num{100000} sentence pairs in \textit{law} domain. The test OOD domain is \textit{blogs}, using data from Convivenza.

We tokenise and truecase all datasets with Moses and use shared BPE with \num{10000} (on IWSLT'14) and \num{32000} (on OPUS and Allegra) for word segmentation~\citep{sennrich-etal-2016-neural}.

\subsection{Models and Evaluation}
All experiments are done with the Nematus toolkit~\citep{sennrich-etal-2017-nematus} based on the Transformer architecture~\citep{NIPS2017_3f5ee243}\footnote{\url{https://github.com/chaojun-wang/progressive-translation}}. The baseline is trained on the training corpus without using intermediate sequences. We follow~\citet{wang-sennrich-2020-exposure} to set hyperparameters (see Appendix) on three datasets. For our framework, we scale up the token batch size proportional to the length of the target for a fair comparison, e.g. if the target-side sequence is three times longer than the original target, we scale up the batch size to three times as well.\footnote{Scaling up the token batch size only brings negligible improvement on the baseline.}. The performance of the original order \textit{(lex)-(ali)-tgt} is used for validation and testing. We conduct early-stopping if the validation performance underperforms the best one over 10 times of validation in both the translation quality (BLEU) and the cross entropy loss.

We also compare to two recently proposed methods of domain robustness in NMT. SSMBA~\citep{ng-etal-2020-ssmba} generates synthetic training data by moving randomly on a data manifold with a pair of corruption and reconstruction functions. Reverse+Mono+Replace~\citep{sanchez-cartagena-etal-2021-rethinking} (RMP)  introduces three auxiliary tasks where the target history is less informative. 

We report cased, detokenised BLEU~\citep{papineni-etal-2002-bleu} with SacreBLEU~\citep{post-2018-call}\footnote{Signature: BLEU|\#:1|c:mixed|e:no|tok:13a|s:exp|v:2.1.0}. Each experiment is independently run for three times, and we report the average and standard deviation to account for optimiser instability.

\begin{table*}
\centering
\begin{small}
\begin{tabular}{lcccccc}
\toprule
& \multicolumn{2}{c}{\textbf{IWSLT'14}} & \multicolumn{2}{c}{\textbf{OPUS}} & \multicolumn{2}{c}{\textbf{DE$\to$RM}} \\
\cmidrule(lr){2-3}\cmidrule(lr){4-5}\cmidrule(lr){6-7}
augmentation & in-domain & average OOD & in-domain & average OOD & in-domain & average OOD\\
\midrule
\multicolumn{7}{l}{\textit{Results reported by~\citet{sanchez-cartagena-etal-2021-rethinking}:}} \\
Transformer & 30.0$_{\pm 0.10}$    & \phantom{0}8.3$_{\pm 0.85}$ & - & - & - & - \\
RMP & 31.4$_{\pm 0.30}$  & 11.8$_{\pm 0.48}$ & - & - & - & - \\
\multicolumn{7}{l}{\textit{Results reported by~\citet{ng-etal-2020-ssmba}:}} \\
Transformer & - & - & 57.0\phantom{$_{\pm 0.15}$}  & 10.2\phantom{$_{\pm 0.15}$} & 51.5\phantom{$_{\pm 0.15}$} & 12.2\phantom{$_{\pm 0.15}$} \\
SSMBA & - & - & 54.9\phantom{$_{\pm 0.15}$} & 10.7\phantom{$_{\pm 0.15}$} & 52.0\phantom{$_{\pm 0.15}$} & 14.7\phantom{$_{\pm 0.15}$} \\
\midrule
\multicolumn{5}{l}{\textit{Our experiments:}} \\
Transformer & 32.1$_{\pm 0.38}$  & 13.9$_{\pm 0.19}$ & 58.8$_{\pm 0.38}$ & 11.0$_{\pm 0.22}$  & 54.4$_{\pm 0.25}$ & 19.2$_{\pm 0.23}$ \\
SSMBA & 31.9$_{\pm 0.15}$  & 15.4$_{\pm 0.10}$ & 58.4$_{\pm 0.20}$ & 12.1$_{\pm 0.21}$ & 54.7$_{\pm 0.20}$ & 20.4$_{\pm 0.15}$ \\
RMP & 32.2$_{\pm 0.06}$  & 14.7$_{\pm 0.17}$ & 59.2$_{\pm 0.25}$  & 12.6$_{\pm 0.41}$ & 55.1$_{\pm 0.21}$ & 21.5$_{\pm 0.23}$ \\
PT$_{simple}$ & 31.2$_{\pm 0.50}$ & 16.1$_{\pm 0.30}$ & 58.5$_{\pm 0.64}$ & 12.1$_{\pm 0.18}$ & 54.6$_{\pm 0.12}$ & 20.3$_{\pm 0.31}$ \\
PT$_{full}$ & 30.5$_{\pm 0.21}$  & 17.0$_{\pm 0.35}$ & 58.4$_{\pm 0.12}$ & 12.6$_{\pm 0.10}$ & 54.4$_{\pm 0.21}$ & 20.4$_{\pm 0.51}$ \\
\bottomrule
\end{tabular}
\end{small}
\captionof{table}{Average BLEU ($\uparrow$) and standard deviation on in-domain and out-of-domain test sets for models trained on IWSLT'14 DE$\to$EN, OPUS DE$\to$EN and Allegra DE$\to$RM. PT$_{simple}$: method \textcircled{2} in Table~\ref{tab:6}; PT$_{full}$: method \textcircled{8} in Table~\ref{tab:6}; RMP: Reverse+Mono+Replace}
\label{tab:3}
\end{table*}

\subsection{Results}
\label{sec:4.4}
We test our proposal mainly on IWSLT'14 DE$\to$EN. Table~\ref{tab:6} summarises the results. \textcircled{1} is the baseline system which is trained on parallel corpus only without any data augmentation. The average OOD is computed by averaging results across all OOD test sets.

\noindent\textbf{Single \textit{lex} benefits OOD whereas \textit{ali} does not.} Firstly, we simply prepend the produced intermediate sequence(s) (any one of them and both of them in the order of \textit{lex}-\textit{ali}) to the target sequence. Results show that single \textit{lex} (\textcircled{2}) significantly improves the OOD performance by 2.2 BLEU, at the cost of 0.9 BLEU decrease in in-domain performance. However, the introduction of \textit{ali} deteriorates the performance on both in-domain (ID) and OOD test sets (\textcircled{3} and \textcircled{4}). We argue that this comes from the reason that the learning of generating \textit{ali} is more difficult than generating \textit{lex} (\textit{ali} needs an extra reordering step and also the produced \textit{ali} is noisy due to the word alignment errors). As a result, \textit{ali} is more erroneous than \textit{lex} during inference. Therefore, the generation quality of the target deteriorates due to its causal dependency on \textit{ali}.

\noindent\textbf{\textit{ali} benefits OOD with the support of permutation multi-task learning.} We try to alleviate the problem by introducing the permutation multi-task learning on top of \textcircled{2}$\sim$\textcircled{4}. Results show that the permutation successfully alleviates the deterioration of introducing  \textit{ali}, bringing positive results for both ID and OOD (\textcircled{3}$\rightarrow$\textcircled{6}, \textcircled{4}$\rightarrow$\textcircled{7}). With the permutation, a single \textit{ali} intermediate sequence (\textcircled{6}) can improve OOD over the baseline by 2 BLEU and the combination of \textit{lex} and \textit{ali} (\textcircled{7}) bring further improvement on OOD over single \textit{lex} (\textcircled{2}) or single \textit{ali} (\textcircled{6}) by 0.5 and 0.7 BLEU respectively. The permutation shows a negative effect on single \textit{lex} (\textcircled{2}$\rightarrow$\textcircled{5}). Because the \textit{lex} is very easy to learn, few error would occur when predicting \textit{lex}. Therefore, permutation is not effective and even has negative effects as it makes the neural model hard to focus on learning the task of \textit{lex}-\textit{tgt}, leading to inferior performance.

\noindent\textbf{MBR decoding brings further improvement.} For the \textit{lex, ali, tgt} with permutation, there are six permutations in total. We dynamically select a \textit{consensus} translation over each input data by performing MBR decoding over translation from all permutations. Results show MBR (\textcircled{7}$\rightarrow$\textcircled{8}) could further improve the OOD and ID performances by 0.4 and 0.6 BLEU respectively, and outperforms baseline OOD by 3.1 BLEU at  the cost of 1.6 BLEU decrease in ID.

\noindent\textbf{Results on other datasets and comparison with existing methods.} As \textcircled{8} achieves the highest OOD performance and \textcircled{2} achieves relatively high OOD and ID performance with simpler techniques, we name \textcircled{8} as PT$_{full}$ and \textcircled{2} as PT$_{simple}$ and evaluate these two methods on another two domain-robustness datasets (OPUS DE$\to$EN and Allegra DE$\to$RM). Table~\ref{tab:3} lists the results.

Baselines (Transformer) in cited works (RMP and SSMBA) are trained under inappropriate hyperparameters, e.g. on IWSLT'14, the cited works uses default hyperparameters for the WMT dataset (more than 10 times larger than IWSLT'14). To enable better comparison by other researchers, we train the Transformer with the appropriate hyperparameters provided by~\citet{wang-sennrich-2020-exposure} to build strong baselines, which outperform those in the cited works. We re-implement the other two DA methods based on our baseline for comparison. 

Results show that both PT$_{simple}$ and PT$_{full}$ perform most effectively on IWSLT'14 OOD, surpassing the existing methods by 0.7-2.3 BLEU. On the other two new datasets, PT$_{simple}$ and PT$_{full}$ show consistent OOD improvement, outperforming our baseline (Transformer) by 1.1-1.6 BLEU and 1.1-1.2 BLEU on OPUS and DE$\to$RM dataset respectively. The ID performance of PT$_{simple}$ and PT$_{full}$ on these two datasets is less affected than on IWSLT'14, at the cost of 0.3-0.4 BLUE decrease on OPUS and even no decrease on the Allegra DE$\to$RM.

PT$_{full}$ significantly outperforms PT$_{simple}$ OOD on OPUS DE$\to$EN and they show negligible ID differences. For Allegra DE$\to$RM, PT$_{simple}$ and PT$_{full}$ shows similar OOD and ID performance.

\section{Analysis}
BLEU score indicates that the proposed methods can improve domain robustness. In this section, we investigate the reduction of hallucinations and performance on larger datasets of our methods.
\subsection{Hallucinations}
\label{sec:4.5}
Hallucinations are more pronounced in out-of-domain translation, and their misleading nature makes
them particularly problematic. 
Therefore, many works have been conducted on hallucinations, involving detection of hallucinations~\citep{zhou-etal-2021-detecting, guerreiro2022looking, dale2022detecting}, exploration of the causes of hallucinations~\citep{raunak-etal-2021-curious, yan2022probing}, and finding solutions for hallucinations~\citep{miao-etal-2021-prevent, muller-sennrich-2021-understanding} etc.

To test our methods for reducing the hallucinations under domain shift, we manually evaluate the proportion of hallucinations on IWSLT'14 and OPUS (DE$\to$EN) OOD test sets. We follow the definition and evaluation by~\citet{muller-etal-2020-domain}, considering a translation as a hallucination if it is \textbf{(partially) fluent} and its content is not related to the source \textbf{(inadequate)}. We report the proportion of such hallucinations in each system.

The manual evaluation is performed by two students who have completed an English-medium university program. We collect $\sim$3000 annotations for  10 configurations. We ask annotators to evaluate translations according to fluency and adequacy. For fluency, the annotator classifies a translation as fluent, partially fluent or not fluent; for adequacy, as adequate, partially adequate or inadequate.
We report the  kappa coefficient (K)~\citep{carletta-1996-assessing} for inter-annotator and intra-annotator agreement in Table~\ref{tab:7}, and assess statistical significance with Fisher’s exact test (two-tailed). 

Table~\ref{tab:8} shows the results of human evaluation. All of the DA methods significantly decrease the proportion of hallucinations by 2\%-6\% on IWSLT'14 and by 9\%-11\% on OPUS, with the increase in BLEU. Note that the two metrics do not correlate perfectly: for example, PT$_{full}$ has a higher BLEU than PT$_{simple}$ but PT$_{simple}$ has a similar or even lower proportion of hallucinations than PT$_{full}$. This indicates that PT$_{full}$ improves  translation quality in other aspects.

\begin{table}
\centering
\begin{small}
\setlength{\tabcolsep}{0.5em}
\begin{tabular}{lcccccc}
\toprule
& \multicolumn{3}{c}{\textbf{inter-annotator}} & \multicolumn{3}{c}{\textbf{intra-annotator}} \\
\cmidrule(lr){2-4}\cmidrule(lr){5-7}
annotation & $P(A)$ & $P(E)$ & $K$ & $P(A)$ & $P(E)$ & $K$ \\
\midrule
fluency & 0.52 & 0.31 & 0.30 & 0.84 & 0.39 & 0.73 \\
adequacy & 0.68 & 0.38 & 0.48 & 0.88 & 0.38 & 0.81 \\
\bottomrule
\end{tabular}
\end{small}
\caption{Inter-annotator (N=300) and intra-annotator agreement (N=150) of manual evaluation.}
\label{tab:7}
\end{table}

\begin{table}[!ht]
\centering
\small
\begin{tabular}{lcc}
\toprule
& \multicolumn{2}{c}{\% hallucinations (BLEU)} \\
\cmidrule{2-3}
Augmentation  & IWSLT'14 & OPUS \\
\midrule
Transformer  & 11\% (13.9) & 39\% (11.0) \\
RMP & \phantom{0}9\% (14.7) & 30\% (12.6) \\
SSMBA & \phantom{0}6\% (15.4) & 28\% (12.1)\\
PT$_{simple}$ & \phantom{0}5\% (16.1) & 28\% (12.1) \\
PT$_{full}$ & \phantom{0}7\% (17.0) & 30\% (12.6) \\
\bottomrule
\end{tabular}
\caption{Proportion of hallucinations ($\downarrow$) and BLEU ($\uparrow$) on out-of-domain test sets over IWSLT'14 and OPUS (DE$\to$EN).}
\label{tab:8}
\end{table}

\subsection{Tendency by scaling up the corpus size}
\label{sec:4.6}
\begin{figure*}[!t]
    \centering
    \includegraphics[width=0.75\linewidth]{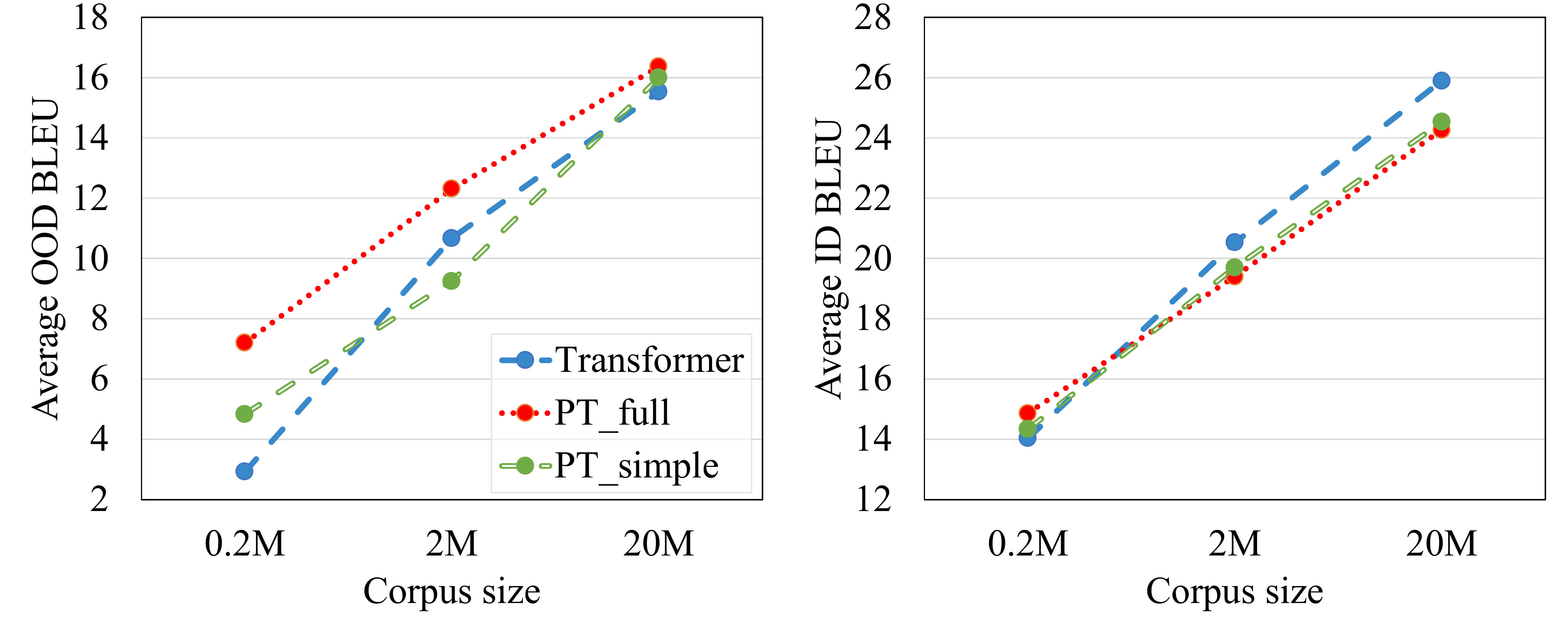}
    \caption{Average BLEU ($\uparrow$) on in-domain and out-of-domain test sets for models trained on OPUS DE$\to$EN (\textit{subtitles}) with various sizes of the training corpus.}
    \label{fig:4}
\end{figure*}
Since the size of the training corpus in the previous experiments ranges from 0.1M to 0.6M (million) samples, which is a low-resource setting for NMT, here we investigate the performance of our methods when scaling up the corpus size.
We use \textit{subtitles} domain from OPUS as the in-domain training data (because it has around 20M sentence pairs) and the rest four domains as the OOD test sets. We use the first 0.2M, 2M and 20M samples in the corpus as the training data separately. We follow the same data preprocessing as for OPUS (\textit{medical}). The hyperparameters for training the model are the same as those for IWSLT'14 when the corpus size is 0.2M and those for OPUS (\textit{medical}) when the corpus size is 2M. For the corpus size of 20M, we increase the token batch size to 16384 instead of 4096 and keep the rest of the hyperparameters the same as for the 2M corpus size. Similarly, each experiment is independently run for three times and we report the average result.

Results are shown in Figure~\ref{fig:4}. As expected, increasing the corpus size (0.2M-20M) improves both ID and OOD performance for all systems. When the corpus size is small (0.2M), PT$_{full}$ (red line) shows a considerable improvement in OOD over the baseline (blue line) by 4.3 BLEU and even slightly benefits ID, surpassing the baseline by around 0.9 BLEU. However, scaling up the corpus size (0.2M-20M) narrows the gap of OOD improvement  (4.3-0.9 BLEU) between the baseline and PT$_{full}$, and widens the ID deterioration from +0.9 to -1.6 BLEU. 

In general, PT$_{simple}$ (green line) follows a similar tendency as PT$_{full}$, compared to the baseline. However, PT$_{simple}$ underperforms the baseline at the corpus size of 2M. By a close inspection, we found that the training of PT$_{simple}$ is relatively unstable. The standard deviations of PT$_{simple}$ for OOD are 1.38, 2.49 and 0.24 on 0.2M, 2M and 20M corpus size respectively, whereas the standard deviations of PT$_{full}$ are 0.47, 0.27 and 0.52 respectively. This indicates that the training of PT$_{simple}$ is less stable than PT$_{full}$ when the corpus size is 0.2M-2M. The  better stability of PT$_{full}$ may come from its  permutation multi-task learning mechanism.

PT$_{simple}$ always underperforms PT$_{full}$ on OOD for any corpus size. PT$_{simple}$ shows slightly better ID performance than PT$_{full}$ when the corpus size is large (2M-20M) but underperforms PT$_{full}$ on ID performance in low resource setting where the corpus size is 0.2M.

\section{Conclusion}
Our results show that our introduced intermediate signals effectively improve the OOD performance of NMT. Intermediate sequence \textit{lex} can benefit OOD by simply prepending it to the target. \textit{ali} is more likely to be erroneous during inference than \textit{lex}, which results in degenerated target due to the spurious causal relationship. Our proposed permutation multi-task learning successfully alleviates the problem and manifests the effectiveness of \textit{ali}. Experiments also confirm that the MBR algorithm can further improve the performance by dynamically selecting a \textit{consensus} translation from all permutations. The human evaluation shows that the proposed methods substantially reduce the number of hallucinations of the out-of-domain translation. Experiments on the larger corpus sizes indicate that our methods are especially promising in the low-resource scenarios.

Our work is the first attempt to complete the puzzle of the study of intermediate signals in NMT, and two new ideas may benefit this study in other areas: 1) thinking intermediate signals from the intermediate structures between the transformation from the input to the output; 2) the permutation multi-task learning, instead of only pre/appending intermediate sequences to the output sequence. The permutation multi-task learning + MBR decoding framework is also a potential solution for any multi-pass generation tasks (e.g. speech translation), which suffer from the error propagation problem. The problem is alleviated with the permutation which disentangles causal relations between intermediate and final results. Finally, our work provides a new perspective of data augmentation in NMT, i.e. augmenting data by introducing extra sequences instead of directly modifying the source or target.

\section{Limitations}
The way we use the intermediate sequences is to concatenate new sequences and the target sequence as the new target. As a result, the length of the target increases linearly with the number of intermediate sequences introduced, which increases the cost of inference. In the meantime, Minimum Bayes Risk decoding needs to do prediction multiple times under different control tasks, which further increases the computational cost. However, there are potential solutions to compromise between the computational cost and quality, e.g. learning a student model by distilling the domain-robust knowledge from Progressive Translation.

\section{Ethics Statement}
The datasets used in the experiments are all well-known machine translation datasets and publicity available. Data preprocessing does not involve any external textual resources. Intermediate sequences generated in our data augmentation method are new symbolic combinations of the tokens in the target language. However, the final output of the model is the \textit{tgt} sequence which is the same as the target sequence in the original training set. Therefore, we would not expect the model trained with our data augmentation method would produce more harmful biases. Finally, we declare that any biases or offensive contexts generated from the model do not reflect the views or values of the authors. 


\bibliography{anthology,custom}
\bibliographystyle{acl_natbib}

\clearpage
\appendix
\section{Appendix}

\subsection{Discussion of Intermediate Sequences}

\textit{lex} and \textit{ali} intermediate sequences may come from certain intermediate topological spaces between the transformation from the topological spaces of the source into the target languages. We empirically confirm that such intermediate sequences might look strange but are easier for the neural model to learn and predict, since they are structurally closer to the source. We use the standard Transformer model to learn to predict \textit{lex}, \textit{ali} and \textit{tgt} (this is just the baseline) directly on IWSLT'14 dataset and report the results on both in-domain and out-of-domain test sets. Note that the gold-standard sequences of \textit{lex} and \textit{ali} on the out-of-domain test sets are produced on the corresponding out-of-domain training sets.

Table~\ref{tab:10} shows that \textit{lex} is easier to be predicted than \textit{ali}, and \textit{ali} is easier to be predicted than \textit{tgt} by the NMT model, over both in-domain and out-of-domain test sets.
\begin{table}[!h]
\centering
\begin{small}
\begin{tabular}{lccc}
\toprule
\textbf{Domain} & \textbf{lex} & \textbf{ali} & \textbf{tgt} \\
\midrule
ID  & 94.0$_{\pm 0.20}$ & 61.1 $_{\pm 0.12}$  & 32.1$_{\pm 0.38}$ \\
OOD & 72.6$_{\pm 0.60}$ & 47.9 $_{\pm 0.48}$ & 13.9$_{\pm 0.19}$ \\
\bottomrule
\end{tabular}
\end{small}
\caption{Average BLEU ($\uparrow$) and standard deviation on in-domain and out-of-domain test sets on IWSLT'14 DE$\to$EN when the target is \textit{lex}, \textit{ali} or \textit{tgt} separately.}
\label{tab:10}
\end{table}

\clearpage
\subsection{Hyperparameters}
\vspace*{\fill}
\noindent\begin{minipage}{\textwidth}
\centering
\begin{tabular}{llr}
\hline
& IWSLT & OPUS/Allegra \\
\cline{2-3}
embedding layer size  & \multicolumn{2}{c}{512} \\
hidden state size & \multicolumn{2}{c}{512} \\
tie encoder decoder embeddings  & \multicolumn{2}{c}{yes} \\
tie decoder embeddings  & \multicolumn{2}{c}{yes} \\
loss function & \multicolumn{2}{c}{\phantom{AAAA}per-token-cross-entropy\phantom{AAAA}} \\
label smoothing & \multicolumn{2}{c}{0.1}\\
optimizer & \multicolumn{2}{c}{adam} \\
learning schedule & \multicolumn{2}{c}{transformer} \\
warmup steps & 4000 & 6000 \\
gradient clipping threshold & 1 & 0 \\
maximum sequence length &\multicolumn{2}{c}{100}\\
token batch size & \multicolumn{2}{c}{4096}\\
length normalization alpha & 0.6 & 1 \\
encoder depth & \multicolumn{2}{c}{6}\\
decoder depth & \multicolumn{2}{c}{6}\\
feed forward num hidden & 1024 & 2048 \\
number of attention heads & 4 & 8 \\
embedding dropout & 0.3 & 0.1 \\
residual dropout & 0.3 & 0.1 \\
relu dropout & 0.3 & 0.1 \\
attention weights dropout & 0.3 & 0.1 \\
beam size & \multicolumn{2}{c}{4}\\
validation frequency & \multicolumn{2}{c}{4000 iterations} \\
\hline
\end{tabular}
\captionof{table}{Configurations of NMT systems over three datasets.}
\label{tab:9}
\end{minipage}
\vspace*{\fill}

\end{document}